\documentclass[sigconf]{acmart}

\copyrightyear{2026}
\acmYear{2026}
\setcopyright{cc}
\setcctype{by-nc-nd}
\acmConference[SIGSPATIAL '26]{The 34th ACM International Conference on Advances in Geographic Information Systems}{November 03--06, 2026}{Riverside, CA, USA}
\acmBooktitle{The 34th ACM International Conference on Advances in Geographic Information Systems (SIGSPATIAL '26), November 03--06, 2026, Riverside, CA, USA}
\acmDOI{10.1145/3841645.3843276}
\acmISBN{979-8-4007-2950-8/2026/11}

\usepackage[utf8]{inputenc} 
\usepackage[T1]{fontenc}    
\usepackage{hyperref}       
\usepackage{url}            
\usepackage{booktabs}       
\usepackage{amsfonts}       
\usepackage{nicefrac}       
\usepackage{microtype}      

\usepackage{diagbox}
\usepackage{amsmath}
\usepackage{float}
\usepackage{graphicx}
\usepackage{subcaption}

\usepackage{xcolor}

\usepackage{pifont}
\newcommand{\cmark}{\ding{51}} 
\newcommand{\xmark}{\ding{55}} 

\usepackage{multirow}
\usepackage{wrapfig}

\usepackage{svg}

\title{Multimodal Road Closure Mapping via Cross-Modal Graph Attention}
\author{Izzeddin Teeti}
\email{izzteeti@amazon.com}
\affiliation{%
  \institution{Amazon}
  \country{US}
}

\author{Vaibhav Kango}
\email{kangov@amazon.com}
\affiliation{%
  \institution{Amazon}
  \country{US}
}

\author{Nidhi Sridhar}
\email{nidhsrid@amazon.com}
\affiliation{%
  \institution{Amazon}
  \country{US}
}

\author{Mohamed Moustafa}
\email{mmoustm@amazon.com}
\affiliation{%
  \institution{Amazon}
  \country{US}
}

\begin{document}

\begin{abstract}
  Accurate and timely identification of road closures is critical for large-scale navigation and logistics systems, yet remains challenging due to the need to jointly reason over visual evidence, spatial context, and road network topology. Existing approaches either rely on manual map updates or focus on detecting closure-related objects without resolving which road segment is affected. In this work, we introduce a unified multimodal and multitask framework that jointly performs \emph{road closure classification} (no, partial, complete) and \emph{road segment association} within a single learning architecture. Our model integrates graph-structured map data, satellite imagery augmented with projected street-level detections, and street-view images using modality-specific encoders and cross-modal transformer attention, enabling explicit reasoning over visual cues and network topology. Evaluated on nearly 20K real-world scenarios collected from a large-scale commercial vehicle fleet, our approach achieves \textbf{72.7\%} closure classification accuracy and \textbf{60\%} road segment association accuracy, outperforming frontier vision--language models by over \textbf{28 percentage points} on closure classification while being over \textbf{4{,}000$\times$ smaller} and \textbf{100$\times$ faster} at inference. Extensive ablations demonstrate the importance of cross-modal attention, edge-aware pooling, and graph message passing for accurate geospatial reasoning. This work establishes the first benchmark for joint road closure classification and segment-level attribution, and shows that task-specific architectural inductive biases substantially outperform general-purpose multimodal models for structured geospatial understanding.

\end{abstract}




\maketitle


\section{Introduction}
\label{sec:intro}

Accurate and timely road network information is fundamental to high-quality digital maps to enable safe and efficient delivery planning and navigation at scale. With billions of packages delivered annually by large-scale logistics operators, temporary road closures caused by construction, accidents, or events are a major source of routing failures. Yet most map systems rely on manual reporting, leading to delayed or incomplete updates that degrade navigation quality and operational efficiency.

Updating maps with road closures is more challenging than handling static or well-defined attributes such as traffic lights or stop signs. While detecting a traffic light unambiguously implies its presence, observing closure-related objects (e.g., barricades, drums, cones) does not necessarily indicate that a road is closed, nor does it specify which road segment is affected. Translating raw sensor detections into actionable map updates therefore requires solving two coupled problems: determining the closure type and identifying the impacted road segment. This is inherently a multimodal reasoning task.

Geospatial context information is often complementary but incomplete across individual modalities. Street-level images capture fine-grained object appearance but lack global spatial context; satellite imagery provides aerial view layout while missing small and especially transient objects; and map data encodes road topology without visual evidence. Effective road closure mapping requires jointly reasoning over visual, spatial, and graph-structured information. Our approach encodes each modality into a shared token space and fuses them through cross-modal attention, enabling road graph tokens to selectively attend to the most relevant visual evidence.

Recent vision-language models (VLMs) exhibit strong general-purpose multimodal capabilities, yet our experiments show that they struggle with precise geospatial reasoning over structured maps. For example, the strongest frontier VLM we evaluate achieves less than 45\% accuracy on closure type classification and cannot meaningfully address road segment association due to its inability to reason over graph-structured spatial representations. These results suggest that general multimodal understanding does not readily transfer to tasks requiring fine-grained spatial reasoning over road networks, motivating architectures with explicit topological inductive biases.

In this work, we formally define the \emph{road closure classification and road segment association problem} and propose a unified multimodal and multitask architecture as solution. To the best of our knowledge, this is the first work to jointly address closure classification and road segment association within a single learning framework. As discussed in Section~\ref{sec:related_work}, prior approaches primarily focus on detecting closure-related objects, without resolving closure type or performing automated map updates.

We argue that effective geospatial reasoning requires explicit architectural inductive biases. Road networks are inherently graph-structured, where connectivity and topology are as critical as visual appearance. Accordingly, we design a model that explicitly encodes road graph structure, spatial relationships, and cross-modal interactions. Our contributions are summarized as follows:
\begin{enumerate}
    \item We formally define the road closure classification and road segment association tasks, providing a clear problem formulation and baseline for future research.
    \item We propose a modular multimodal, multitask architecture that jointly performs closure classification and segment association.
    \item We extend GATv2 \cite{brody2021attentive} with an edge-aware decoder that conditions predictions on fused multimodal context and supports variable-length edge sets. On a large-scale real-world dataset, our model achieves 72.7\% closure classification accuracy and 60\% segment association accuracy, outperforming frontier VLMs by over 28\% on closure classification while being over 4{,}000$\times$ smaller and 100$\times$ faster.
\end{enumerate}

\section{Related Work}
\label{sec:related_work}

\subsection{Map Association of Detected Road Assets}
Several works study the association of detected traffic-related objects with map elements, but not in the context of road closures or edge-level attribution. Zhang et al.~\cite{zhang2021automated} and Campbell et al.~\cite{Campbell2019} address the association of detected objects, such as stop signs and traffic lights, to digital maps. While related in spirit, these approaches differ fundamentally from our setting. Zhang et al. combine learning-based detection with model- and rule-driven association, but do not report quantitative benchmarks or release code, limiting reproducibility and scalability. Campbell et al. localize traffic signs using street-view imagery alone, employing map data only during the association stage. This reliance on a single sensing modality limits robustness and fails to exploit complementary cues from aerial imagery and map topology, as demonstrated in our ablation study in Section~\ref{sec:ablation}. Importantly, neither work addresses road closures, temporary changes, or edge-level reasoning over road graphs, which are central to our problem setting, nor do they provide publicly available implementations.

\subsection{Road Closure Detection and Roadwork Understanding}
Only a limited number of studies explicitly address road closures, primarily from a detection perspective. Hasan et al.~\cite{Hasan2022} propose a system for detecting road closure signs using YOLO-based object detection~\cite{redmon2016you}. However, their work focuses solely on sign detection and does not describe how detected closures are integrated into a digital map, nor how ambiguities are resolved in complex scenarios such as intersections or multi-lane roads. Moreover, no quantitative evaluation is provided for closure-to-map integration accuracy. More recent datasets such as ROADWork~\cite{ghosh2024roadwork} and methods for construction-site detection~\cite{feng2025robust} advance visual understanding of roadwork scenes. While these efforts improve object-level perception, they do not address closure type classification, road segment association, or reasoning over structured road networks.

\subsection{Graph-Based Reasoning on Road Networks}
Graph neural networks (GNNs) have been widely applied to road networks for tasks such as traffic prediction and road attribute inference~\cite{he2020roadtagger,Zhang2023,Remmouche2025}. These works demonstrate that modeling road networks as graphs provides a strong inductive bias for reasoning over structured, topologically constrained data and enables effective propagation of contextual information across connected road segments. This motivates our use of GNNs for edge-level road closure reasoning. However, existing approaches do not consider closure attribution, multimodal fusion with visual cues, or the estimation of temporary road states, which are central to our problem.

In summary, prior work either focuses on detecting roadwork or closure-related objects without associating them to a digital map, or performs map association using rule-based heuristics. To the best of our knowledge, no existing method jointly integrates multimodal visual and map cues with explicit graph-based reasoning to perform edge-level road closure attribution together with closure type classification on an OSM graph.

\section{Proposed Solution}
\label{sec:method}

\subsection{Problem Definition}
\label{sec:problem_def}

Given a local road network graph $\mathcal{G} = (\mathcal{V}, \mathcal{E})$, a satellite image $\mathbf{I}_{\text{sat}}$ augmented with projected street-level detections, and a street-view image $\mathbf{I}_{\text{street}}$, we address two interconnected problems:

\begin{enumerate}
    \item \textbf{Closure Classification:} Determine whether a road closure exists and, if present, classify its type as either \textit{partial} (vehicle can proceed with restrictions) or \textit{complete} (vehicle cannot proceed). This is formulated as a multi-class classification over $C{=}3$ classes: $\hat{y}_{\text{closure}} \in \{\text{no}, \text{partial}, \text{complete}\}$.

    \item \textbf{Road Segment Association:} Identify which specific road segment is impacted by the detected closure. This is formulated as a classification problem over the set of $M$ candidate edges in the local road network graph: $\hat{e} \in \mathcal{E}$.
\end{enumerate}

Both tasks are solved jointly within a single forward pass, enabling shared representation learning across closure type and spatial attribution.


\begin{figure*}
    \centering
    \includegraphics[width=1\linewidth]{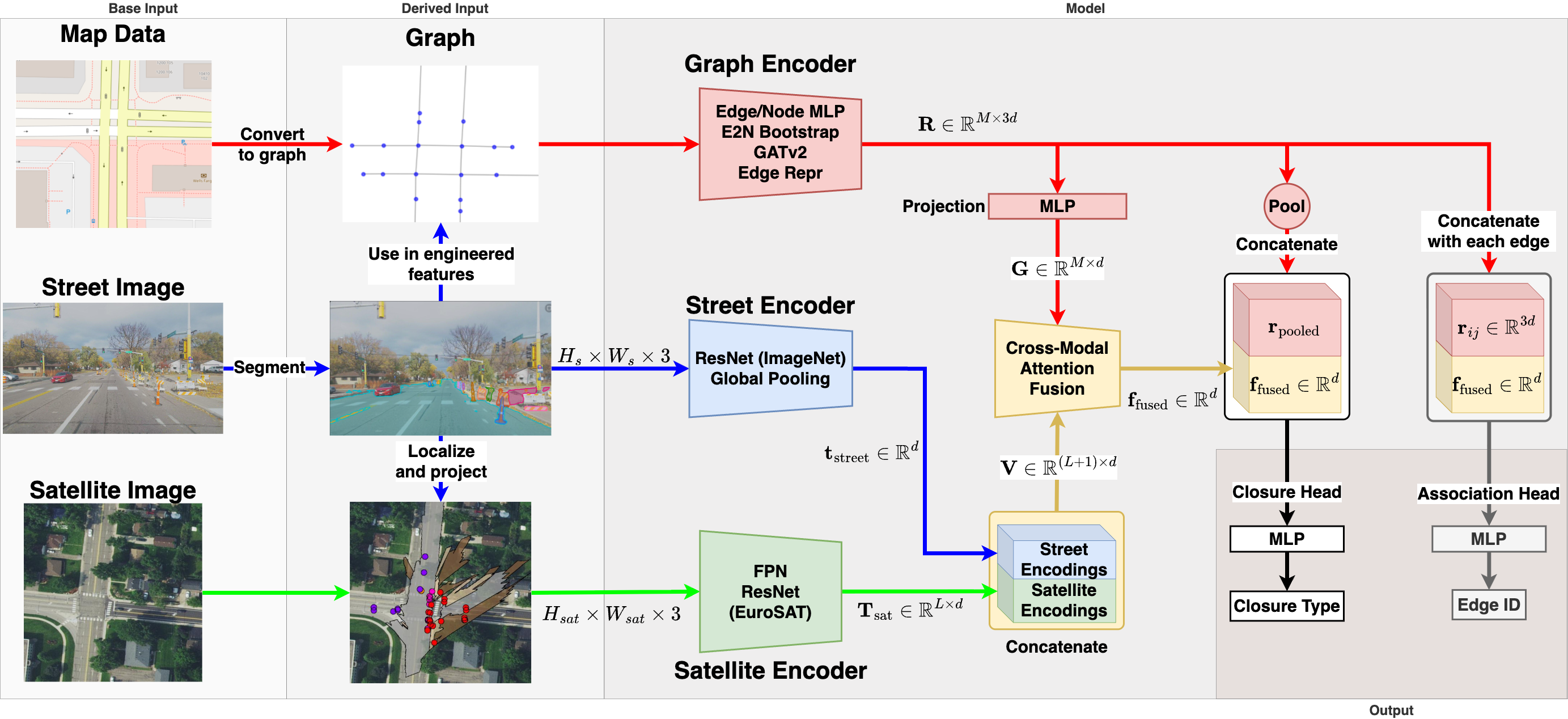}
    \caption{Overview of the proposed multimodal architecture.}
    \label{fig:model_arch}
\end{figure*}

We propose a modular multimodal architecture that encodes each input modality independently, fuses them through cross-modal attention, and produces predictions via task-specific decoder heads. As illustrated in Figure~\ref{fig:model_arch}, the model takes three inputs, street-view images, satellite imagery augmented with projected detections, and map data, and jointly outputs closure classification and impacted road segment association.

\subsection{Modality-Specific Encoders}

\paragraph{Street Image Encoder.}
Street-view images $\mathbf{I}_{\text{street}} \in \mathbb{R}^{H_{st} \times W_{st} \times 3}$ resemble natural images and primarily convey the presence and appearance of closure-related objects. We therefore employ a ResNet backbone pretrained on ImageNet~\cite{ImageNet} to extract a global feature vector $\mathbf{t}_{\text{street}} \in \mathbb{R}^{d}$, where $d$ denotes the hidden dimension.

\paragraph{Satellite Image Encoder.}
Satellite imagery provides a bird’s-eye view of the road network and is augmented with projected street-level detections to encode spatial context. Detected closure objects and road surfaces from street images are geometrically projected onto the satellite view, and the corresponding pixels are replaced with color-coded representations (e.g., barricades in red, signs in blue, road surfaces in gray), as shown in Figure~\ref{fig:satellite_projection}. This results in a three-channel image $\mathbf{I}_{\text{satellite}} \in \mathbb{R}^{H_s \times W_s \times 3}$ that preserves the RGB format while explicitly highlighting detected elements.

This projection introduces a spatial prior that guides the model toward relevant spatial relationships between closure cues and the road layout, without requiring the satellite encoder to relearn object detection. To capture spatial structure at multiple scales, we employ a Feature Pyramid Network (FPN)~\cite{fpn2017} with a ResNet backbone pretrained on satellite imagery~\cite{helber2019eurosat} to better align the feature extractor with the target domain. The FPN produces a set of multi-scale tokens $\mathbf{T}_{\text{satellite}} \in \mathbb{R}^{L \times d}$, where $L$ denotes the total number of spatial locations across pyramid levels.

\paragraph{Design Rationale.}
We intentionally adopt asymmetric encoders to reflect the differing information content of street and satellite imagery. Satellite images encode rich spatial structure and road topology, making multi-scale feature extraction essential. In contrast, street images are perspective views centered on objects ahead of the vehicle, where global appearance cues are sufficient. This asymmetric design balances expressiveness and efficiency, enabling effective multimodal fusion while maintaining practical computational cost.

\paragraph{Graph Encoder.}
The graph encoder transforms raw map data into expressive \emph{edge-level} representations suitable for road segment association. We adopt a bidirectional edge--node--edge design in which edge features initialize node representations, nodes exchange information through message passing to capture topology, and enriched node embeddings are used to refine edge representations. This ensures that each edge captures both its local attributes and its topological context within the road network.

\textbf{Edge and Node Initialization.}
Raw edge attributes (e.g., length, angle, road class) are first encoded using a two-layer MLP with batch normalization to obtain edge embeddings $\mathbf{e}_{ij} \in \mathbb{R}^{d}$. Node features (e.g., intersection properties and spatial cues) are linearly projected to the same hidden dimension. To bootstrap node representations with local road context, we aggregate incoming edge features and combine them with node attributes. Given raw node features $\mathbf{x}_i^{\text{node}}$ and encoded edge features $\mathbf{e}_{ji}$ from incoming edges $j \in \mathcal{N}_{\text{in}}(i)$, the initial node embedding is
\begin{equation}
\mathbf{h}_i^{(0)} = \mathbf{W}_{\text{node}} \mathbf{x}_i^{\text{node}} 
+ \mathbf{W}_{\text{e2n}} \left( \frac{1}{|\mathcal{N}_{\text{in}}(i)|} \sum_{j \in \mathcal{N}_{\text{in}}(i)} \mathbf{e}_{ji} \right),
\label{eq:edge_bootstrap}
\end{equation}
where $\mathbf{W}_{\text{node}}$ and $\mathbf{W}_{\text{e2n}}$ are learnable linear projections. This initialization allows node embeddings to encode both intersection-level attributes and the characteristics of connected road segments prior to message passing.

\textbf{Topology-Aware Message Passing.}
Node embeddings are refined through multiple graph attention layers to propagate information across the road network. We employ GATv2~\cite{brody2021attentive}, which improves expressiveness over GAT~\cite{gat2017}, and extend it to incorporate edge features into the attention computation. Specifically, the unnormalized attention score between nodes $i$ and $j$ is computed as:
\begin{equation}
\alpha_{ij} = \mathbf{a}^\top \text{LeakyReLU}\left( \mathbf{W} [\mathbf{h}_i \,\|\, \mathbf{h}_j \,\|\, \mathbf{e}_{ij}] \right),
\label{eq:gatv2}
\end{equation}
where $\mathbf{h}_i, \mathbf{h}_j$ are node embeddings, $\mathbf{e}_{ij}$ is the edge embedding, and $\mathbf{a}$, $\mathbf{W}$ are learnable parameters. By applying the nonlinearity after concatenation, GATv2 enables dynamic attention rankings that depend on both query and key, unlike standard GAT. This allows message passing to account for node properties and the characteristics of the connecting road segments.

The normalized attention coefficients are obtained via a softmax over the neighborhood of node $i$:
\begin{equation}
\tilde{\alpha}_{ij} = \text{softmax}_j(\alpha_{ij}).
\end{equation}
Node representations are then updated by aggregating messages from neighboring nodes:
\begin{equation}
\mathbf{h}_i^{(\ell+1)} = \sum_{j \in \mathcal{N}(i)} \tilde{\alpha}_{ij} \mathbf{W}_v \mathbf{h}_j^{(\ell)}.
\end{equation}
Residual connections and layer normalization are applied to stabilize training:
\begin{equation}
\mathbf{h}_i^{(\ell+1)} = \text{LayerNorm}\left( \mathbf{h}_i^{(\ell)} + \text{ELU}(\mathbf{h}_i^{(\ell+1)}) \right).
\end{equation}

\textbf{Edge Representation Construction.}
After message passing, the final representation of each road segment is constructed by concatenating the embeddings of its source node, target node, and edge:
\begin{equation}
\mathbf{r}_{ij} = [\mathbf{h}_i \,\|\, \mathbf{h}_j \,\|\, \mathbf{e}_{ij}] \in \mathbb{R}^{3d}.
\end{equation}

Edge representations are padded or truncated to a fixed maximum size $M$ and accompanied by binary masks, yielding a batched tensor $\mathbf{R} \in \mathbb{R}^{B \times M \times 3d}$ for downstream fusion.

\subsection{Cross-Modal Fusion}
\label{sec:fusion}

We fuse graph and visual information using cross-modal attention, where graph edge tokens query visual tokens from the satellite and street encoders. This design allows each road segment to attend selectively to relevant visual evidence.

Edge representations are first projected to the hidden dimension, while visual tokens from satellite imagery and the street-view encoder are concatenated into a shared set of visual tokens (batched as $\mathbf{T}_{\text{satellite}} \in \mathbb{R}^{B \times L \times d}$ and $\mathbf{t}_{\text{street}} \in \mathbb{R}^{B \times 1 \times d}$). To obtain a unified scene representation, we prepend a learnable CLS token to the graph tokens. 

This design follows prior uses of CLS tokens in BERT~\cite{devlin2019bert} for sentence classification, Vision Transformers~\cite{vit2021} for image classification, and CLIP~\cite{clip2021} for vision-language alignment, and is adapted here for graph--vision multimodal fusion. The CLS token acts as an aggregation mechanism that learns to summarize the most relevant multimodal information required for downstream prediction.

Concretely, the projected edge representations $\mathbf{G} \in \mathbb{R}^{M \times d}$ serve as queries, while the concatenated visual tokens $\mathbf{V} \in \mathbb{R}^{(L+1) \times d}$ form keys and values. Prepending the learnable CLS token yields the initial query set
\begin{equation}
\mathbf{Q}^{(0)} = [\mathbf{cls} \,\|\, \mathbf{G}] \in \mathbb{R}^{(1+M) \times d},
\end{equation}
and each cross-attention layer $\ell$ computes
\begin{equation}
\mathbf{Q}^{(\ell+1)} = \text{LayerNorm}\left(
\mathbf{Q}^{(\ell)} + \text{MHA}(\mathbf{Q}^{(\ell)}, \mathbf{V}, \mathbf{V})
\right),
\end{equation}
followed by a position-wise feed-forward network:
\begin{equation}
\mathbf{Q}^{(\ell+1)} = \text{LayerNorm}\left(
\mathbf{Q}^{(\ell+1)} + \text{FFN}(\mathbf{Q}^{(\ell+1)})
\right).
\end{equation}
Here MHA denotes multi-head attention~\cite{vaswani2017attention} with $H$ heads, and FFN is a two-layer feed-forward network with GELU activation. This is stacked over multiple transformer blocks, enabling graph tokens to attend selectively to relevant visual evidence while the CLS token aggregates global multimodal context.

The final fused scene representation is obtained from the CLS token after the last transformer layer and is denoted by $\mathbf{f}_{\text{fused}} \in \mathbb{R}^{d}$. This representation aggregates information across graph and visual modalities and conditions both downstream prediction heads.

\subsection{Task-Specific Decoder Heads}

\paragraph{Association Head.}
The association head predicts which road segment (graph edge) is affected by the closure. Each edge representation is conditioned on the fused scene context by concatenation:
\begin{equation}
\mathbf{z}_{ij} = [\mathbf{r}_{ij} \,\|\, \mathbf{f}_{\text{fused}}] \in \mathbb{R}^{4d}.
\end{equation}
A two-layer MLP produces a scalar logit per edge:
\begin{equation}
s_{ij} = \text{MLP}_{\text{assoc}}(\mathbf{z}_{ij}) \in \mathbb{R}.
\end{equation}
During training, logits corresponding to padded edges are masked by setting them to $-\infty$ prior to softmax. At inference, the predicted road segment is selected as $\hat{e} = \arg\max_j \, \text{softmax}(\mathbf{s})_j$.

\paragraph{Closure Classification Head.}
Closure type classification requires a global representation that respects the variable number of valid edges per graph. We therefore apply edge-aware mean pooling over valid edges only:
\begin{equation}
\mathbf{r}_{\text{pooled}} =
\frac{1}{|\{(i,j)\!:\!m_{ij}=1\}|}
\sum_{(i,j): m_{ij}=1} \mathbf{r}_{ij},
\end{equation}
where $m_{ij}=1$ denotes valid (non-padded) edges. This avoids representation dilution caused by naive pooling over padded edges, as confirmed in our ablation study (Section~\ref{sec:ablation}). The pooled representation is concatenated with the fused context and passed to a classification MLP:
\begin{equation}
\mathbf{y}_{\text{closure}} =
\text{MLP}_{\text{closure}}([\mathbf{r}_{\text{pooled}} \,\|\, \mathbf{f}_{\text{fused}}])
\in \mathbb{R}^{C},
\end{equation}
where $C$ denotes the number of closure classes (no, partial, full).

\subsection{Learning Objective}

We train the model using a multi-task objective with two cross-entropy losses: one for edge association over $M$ candidates and one for closure classification over $C$ classes. The total loss is given by
\begin{equation}
\mathcal{L}
=
\lambda_{\text{assoc}}
\left(-\sum_{j=1}^{M} y_j \log \hat{y}_j \right)
+
\lambda_{\text{closure}}
\left(-\sum_{c=1}^{C} y_c \log \hat{y}_c \right),
\end{equation}
where $y_j$ and $y_c$ are one-hot ground-truth labels,
$\hat{y}_j = \text{softmax}(\mathbf{s})_j$ and
$\hat{y}_c = \text{softmax}(\mathbf{y}_{\text{closure}})_c$
are the predicted probabilities for association and closure classification, respectively, and
$\lambda_{\text{assoc}}$ and $\lambda_{\text{closure}}$ balance the two objectives.

Association is optimized for all samples, including those without closures, as closure-related objects may be present without inducing an actual road closure.

\section{Experiments}
\label{sec:exp}

\subsection{Dataset}
\label{sec:dataset}

We collected a real-world dataset using a large-scale commercial vehicle fleet, comprising 19,380 samples with detected closure objects such as barricades, drums, and cones. The dataset spans diverse geographic contexts across the US, including dense urban areas, suburban neighborhoods, rural highways, and residential streets. Each sample jointly provides street-level imagery, satellite imagery, structured map data, detected closure objects, and precise vehicle pose information (location and heading), enabling the study of multimodal road closure classification and road segment association.

\begin{figure*}[ht]
  \centering
  \begin{subfigure}{0.26\linewidth}
    \centering
    \includegraphics[width=\linewidth]{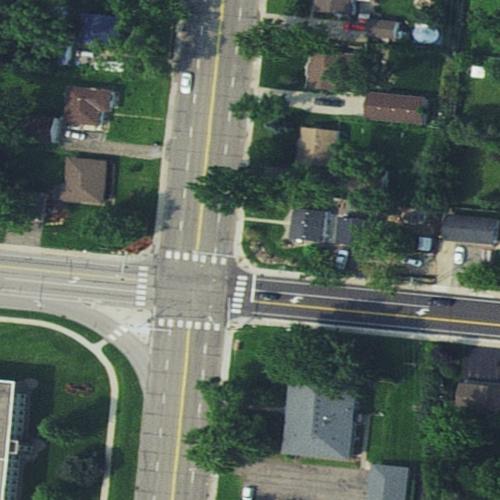}
    \caption{Satellite imagery sample (no projections).}
    \label{fig:satellite_sample}
  \end{subfigure}
  \hfill
  \begin{subfigure}{0.26\linewidth}
    \centering
    \includegraphics[width=\linewidth]{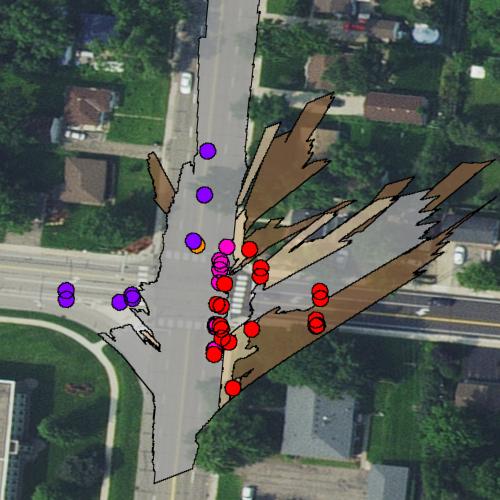}
    \caption{Satellite imagery sample (with projections. Red: barricades, pink: cones/vertical rods, purple: drums, gray: road surfaces).}
    \label{fig:satellite_projection}
  \end{subfigure}
  \hfill
  \begin{subfigure}{0.43\linewidth}
    \centering
    \includegraphics[width=\linewidth]{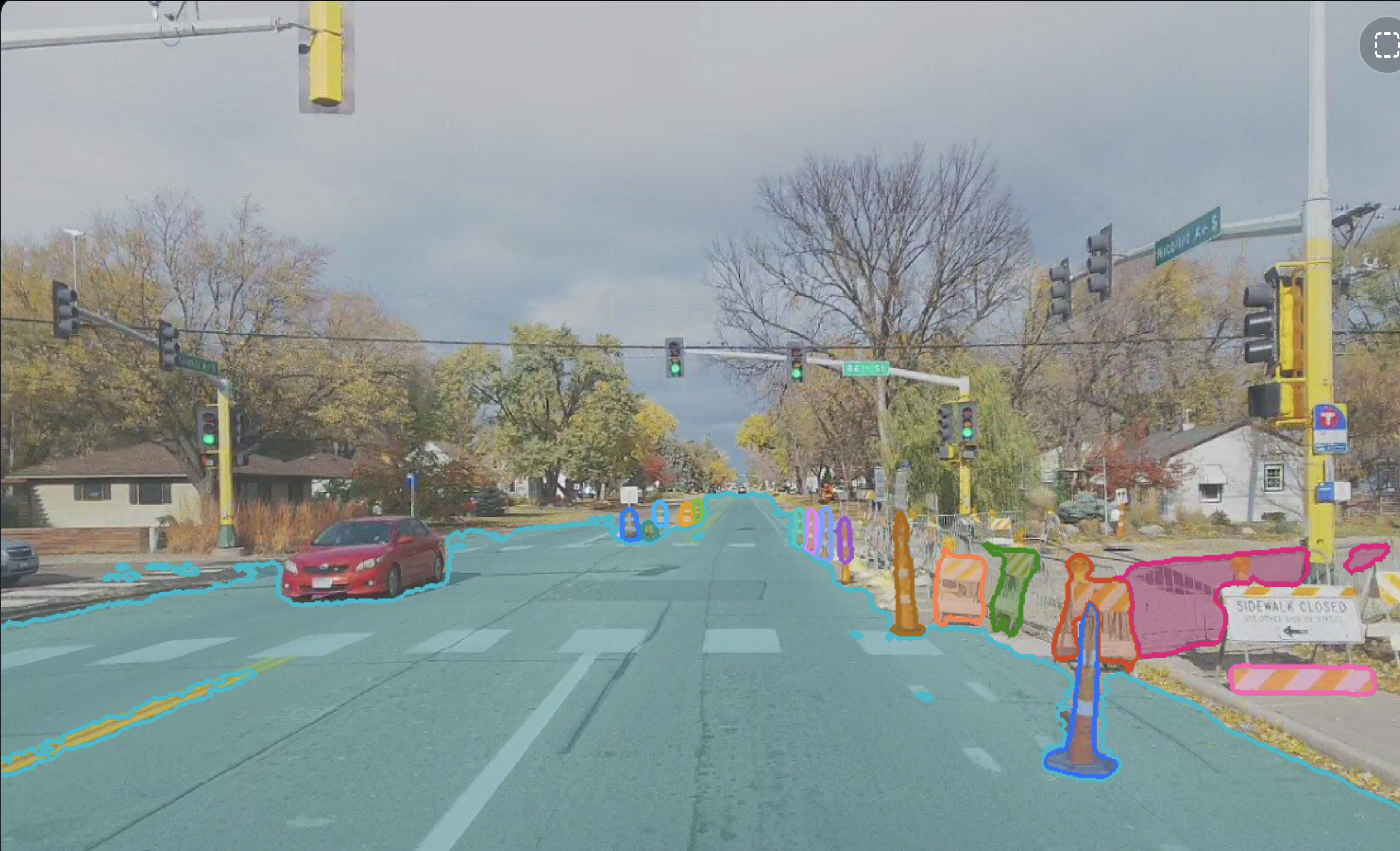}
    \caption{Segmentation of closure objects and surfaces from street imagery.}
    \label{fig:segmentation_sample}
  \end{subfigure}
  \caption{Input modalities for one data sample.}
  \label{fig:satellite_and_projection}
\end{figure*}

Each sample consists of three modalities: (i) a forward-facing street-level image captured by the delivery vehicle (Figure~\ref{fig:segmentation_sample}), (ii) a $100\,\mathrm{m} \times 100\,\mathrm{m}$ satellite image centered at the vehicle location with projected closure objects and surfaces (Figure~\ref{fig:satellite_projection}), and (iii) a local road network graph extracted from map data in the vicinity of the detected object. Data preprocessing, localization, and feature construction are described in Section~\ref{sec:modalities}.

Ground-truth annotations are provided by map and road-network domain experts. Each sample is labeled with the affected road segment and a closure type (\emph{No Closure}, \emph{Partial Closure}, or \emph{Complete Closure}), with a naturally imbalanced distribution of 62\%, 29\%, and 9\% respectively. We split the dataset into training (70\%, 13,566 samples), validation (15\%, 2,907 samples), and test (15\%, 2,907 samples) sets, stratified by closure type.

Our graph encoder operates on a fixed maximum of $M{=}10$ edges per graph. In our dataset, the ground-truth road segment falls within the 10 nearest edges to the vehicle position in 94\% of cases. Samples where the target edge exceeds $M$ are excluded during preprocessing (approximately 6\% of the data). Sequence bucketing or dynamic $M$ selection are noted as potential future optimizations.

\subsection{Data Modalities and Preparation}
\label{sec:modalities}

The system integrates three complementary data modalities. We note that our contribution lies in the multimodal fusion architecture and learning framework for road closure classification and association, rather than in the data acquisition and preprocessing pipeline. Therefore, we treat segmentation, localization, and projection as preprocessing steps performed by existing methods, and focus our technical description on how these modalities are leveraged within our model.

\paragraph{Street-Level Imagery.} Street images are captured via cameras mounted on fleet vehicles. These images undergo semantic segmentation to identify closure objects (arrow boards, barricades, cones, drums, TTC signs, vertical bars, and work vehicles) and surfaces of interest (road, sidewalk, and earth), as illustrated in Figure~\ref{fig:segmentation_sample}. The road closure object categories are adapted from the ROADWork dataset~\cite{ghosh2024roadwork}. Detected objects are localized in 3D using a pinhole camera model with structure-from-motion depth estimates, then projected to absolute geographic coordinates (latitude, longitude) via the vehicle's GPS position and heading. GPS uncertainty (typically $\pm$3--5\,m in urban environments) introduces localization noise; however, the 100\,m $\times$ 100\,m satellite crop used for aerial context provides sufficient spatial margin to accommodate this error without displacing projected objects outside the relevant image region.

\paragraph{Aerial Imagery.}
Based on the GPS coordinates where the vehicle detected a closure object, we retrieve a satellite image $\mathbf{I}_{\text{sat}} \in \mathbb{R}^{H_s \times W_s \times 3}$ covering a $100m \times 100m$ area centered on the vehicle location, as shown in Figure~\ref{fig:satellite_sample}. While not real-time, these images provide crucial spatial context.

\paragraph{Graph-Structured Map Data.} Map data (streets and intersections with attributes such as speed limits and road types, similar to OpenStreetMap format) is converted into a bidirectional graph $\mathcal{G} = (\mathcal{V}, \mathcal{E})$, where roads are edges and intersections are nodes. For each edge we engineer the road class (discrete values 10--102 indicating road type, e.g.\ service road, highway, urban street), the Euclidean distances from the edge's closest point to the detected closure object and to the delivery vehicle, the edge orientation angle relative to true north, and the angle between the edge direction and the vehicle heading. For each node we engineer its $(x, y)$ coordinates relative to the delivery vehicle position, its degree (number of connected edges), the Euclidean distances to the detected closure object and to the delivery vehicle, and the angles of the node-to-object and node-to-vehicle lines relative to true north.

These engineered features collectively encode three key aspects of the scene: the \textit{road network structure} (road class, node degree, edge orientation), the \textit{vehicle state} (distances and angles relative to vehicle position and heading), and the \textit{closure object location} (distances and angles relative to detected objects). This representation enables the model to reason about spatial relationships between the road topology, the vehicle's position, and the detected closure. All distance features use Euclidean distance, which closely approximates the geodesic (haversine) distance for the short ranges encountered in our application (typically within 100m). Feature normalization is applied as follows: distance features are min-max normalized to $[0, 1]$, angular features are encoded as $(\sin\theta, \cos\theta)$ pairs to handle periodicity, and road class undergoes ordinal normalization where the lowest value (10, representing highways) maps to 0 and the highest value (102, representing parking aisles) maps to 1. This ordinal encoding preserves the inherent hierarchical relationship between road types, reflecting their relative importance in the road network. To manage computational complexity while maintaining task-relevant information, we limit the graph to a maximum of $M$ edges nearest to the vehicle location. This design choice is motivated by the observation that road closures typically affect segments in the immediate vicinity of the vehicle; thus, the impacted edge is almost always among the closest edges, making distant edges unnecessary for accurate prediction.

\subsection{Implementation Details}
\label{sec:implementation}

All models are implemented in PyTorch and trained on a single NVIDIA Tesla V100 (16GB). Full architectural specifications and hyperparameters are provided in Appendix~\ref{appendix:hyperparameters}.

\subsection{Baselines}
\label{sec:baselines}

To our knowledge, this work is the first to jointly address road closure classification and affected road segment association using multimodal learning. With no public established benchmarks for this problem, we construct the following baselines:

\begin{enumerate}
    \item \textbf{VLM-A.} A frontier proprietary vision-language model ($>$100B parameters) accessed through an API, evaluated on the closure classification task using street-level imagery.\footnote{We report the two VLM baselines anonymously, as VLM-A and VLM-B; their specific identities are not essential to our findings. Both are widely used general-purpose models at the frontier of publicly available multimodal capability at the time of evaluation. Reported latencies are end-to-end and include API round-trip time.} The model struggles with explicit spatial and topological reasoning over road networks.

    \item \textbf{VLM-B.} A more recent and more capable frontier proprietary VLM ($>$100B parameters), with improved multi-step reasoning and vision capabilities. As with VLM-A, it is evaluated only on closure classification, as general-purpose VLMs fundamentally lack the ability to reason over graph-structured spatial representations required for road segment association.

    \item \textbf{Nearest-Edge Heuristic.} A rule-based baseline that assigns the closure to the road segment nearest to the median of the detected objects. This establishes a lower bound on task difficulty and isolates the contribution of learned representations over simple proximity-based reasoning.

    \item \textbf{Task-Specific Variants.} To assess the benefit of joint learning, we train two single-task variants of our architecture: one optimized solely for closure classification and another solely for road segment association. These models serve as internal baselines in our ablations.
\end{enumerate}

\subsection{Evaluation Metrics}
\label{sec:metrics}

For baseline comparisons (Table~\ref{tab:baselines}), we report overall accuracy for each task, along with inference latency. For ablation studies (Tables~\ref{tab:modality_ablation}--\ref{tab:ablation}), we report Precision, Recall, and $F_1$ score, macro-averaged across classes and tasks. These complementary metrics provide deeper insight into model behavior under class imbalance and reveal performance trade-offs not captured by accuracy alone.

\section{Results}
\label{sec:results}

\subsection{Quantitative Results}

Table~\ref{tab:baselines} compares our proposed model against the baselines. General-purpose vision–language models (VLM-A and VLM-B) are reported only for closure classification, as they cannot meaningfully address the association task due to their inability to reason over graph-structured road network representations. Our joint model achieves \textbf{72.7\%} accuracy on closure classification and \textbf{59.9\%} accuracy on segment association, substantially outperforming all baselines while operating with orders-of-magnitude lower latency and model size.

\begin{table}[h]
\centering
\caption{Performance comparison with baseline methods}
\label{tab:baselines}
\resizebox{0.47\textwidth}{!}{%
\begin{tabular}{lcc}
\toprule
\textbf{Model (params)} & \textbf{Closure/Assoc. Acc. (\%)} & \textbf{Latency (s)} \\
\midrule
Nearest-Edge Heuristic (0) & --- / 45.4 & --- \\
VLM-A ($>$100B) & 34.7 / --- & 7.0 \\
VLM-B ($>$100B) & 44.3 / --- & 5.4 \\
\midrule
Ours (Separate) (12M each) & 71.6 / 58.3 & 0.06 \\
\textbf{Ours (Joint)} (23M) & \textbf{72.7} / \textbf{59.9} & \textbf{0.05} \\
\bottomrule
\end{tabular}%
}
\end{table}

To quantify uncertainty, we compute 95\% bootstrap confidence intervals (10{,}000 resamples) on the test set. Our joint model achieves 72.7\% closure classification accuracy (95\% CI: [71.1\%, 74.4\%]) and 59.9\% segment association accuracy (95\% CI: [58.1\%, 61.7\%]). The narrow confidence intervals ($\leq$3.6 percentage points) indicate stable performance estimates given the test set size of 2{,}907 samples.

Several key observations emerge from these results. First, large general-purpose VLMs struggle with structured geospatial reasoning despite prompt tuning and strong general reasoning capabilities, indicating poor transfer to map-centric tasks. Even the stronger of the two (VLM-B) achieves only 44.3\% on closure classification and cannot meaningfully address road segment association, confirming that general-purpose multimodal understanding does not substitute for explicit architectural inductive biases for graph reasoning and spatial alignment. Second, the nearest-edge heuristic achieves 45.4\% on association through proximity alone, establishing a non-trivial lower bound; our model's +14.5 point improvement over this heuristic confirms that learned multimodal representations contribute substantially beyond simple spatial proximity.

Joint training yields consistent performance gains (+1.1\% for closure classification and +1.6\% for association) over separate models, validating the effectiveness of multi-task learning. Closure type and affected road segment are geometrically coupled: complete closures typically block the full road width, whereas partial closures affect only a subset of the roadway.

\paragraph{Practical Considerations.}
In a potential application scenario, false positives (incorrectly predicting a closure) would cause unnecessary rerouting, while false negatives (missing a real closure) would result in vehicles encountering blocked roads. The asymmetric cost of these errors motivates operating the model as a confidence-thresholded filter: only predictions exceeding a confidence threshold would be acted upon automatically, while lower-confidence cases can be deferred to human review or additional verification passes. This design allows practitioners to tune the precision--coverage tradeoff based on application requirements without retraining the model.

Finally, our approach offers substantial efficiency advantages. The joint model is over $100\times$ faster than VLM-based approaches at inference time while being over $4{,}000\times$ smaller (23M vs.\ $>$100B parameters). This efficiency makes the model suitable for latency-sensitive applications where road closures must be detected and propagated rapidly.  Qualitative results and illustrative anecdotes are presented in Section~\ref{sec:qualitative}.

\subsection{Ablation Studies}
\label{sec:ablation}

Table~\ref{tab:modality_ablation} summarizes a systematic ablation over the input modalities used by our model. Table~\ref{tab:ablation} reports ablations on fusion strategy, graph architecture, reasoning depth, and message passing.

\paragraph{Input Modalities.}
The graph-only baseline achieves a non-trivial performance ($F_1 = 35.9\%$, averaged across both tasks), indicating that engineered, vehicle-aware spatial features encode meaningful information for road closure reasoning. However, this modality alone is insufficient.

\begin{table}[h]
\centering
\caption{Ablation study on input modalities.}
\label{tab:modality_ablation}
\resizebox{0.47\textwidth}{!}{%
\begin{tabular}{cccccc}
\toprule
\textbf{Sat.} & \textbf{Sat.+Proj.} & \textbf{Graph} & \textbf{Street} & \textbf{P / R / F1 (\%)} & \textbf{\# Params} \\
\midrule
\xmark & \xmark & \cmark & \xmark & 39.4 / 32.9 / 35.9 & 218K \\
\cmark & \xmark & \xmark & \xmark & 22.6 / 26.9 / 24.6 & 12.32M \\
\cmark & \cmark & \xmark & \xmark & 34.0 / 30.6 / 32.2 & 12.32M \\
\cmark & \cmark & \cmark & \xmark & 52.9 / \textbf{38.0} / 44.2 & 12.37M \\
\cmark & \cmark & \cmark & \cmark & \textbf{58.0} / 37.5 / \textbf{45.5} & 23.42M \\
\bottomrule
\end{tabular}%
}
\end{table}

Incorporating projection of detected closure objects onto satellite imagery substantially improves performance ($24.6\% \rightarrow 32.2\%$), highlighting the importance of explicitly encoding semantic spatial context. The projection operation introduces a strong spatial prior by localizing closure-related objects within the aerial view, allowing the model to focus on spatial reasoning rather than implicit object localization.

Combining projected satellite imagery with the road network graph yields a further improvement ($F_1 = 44.2\%$), demonstrating the complementarity between visual spatial cues and topological structure. Finally, adding street-level imagery provides a modest but consistent gain ($F_1 = 45.5\%$), capturing fine-grained appearance and perspective information unavailable in overhead views. The street encoder contributes a single CLS token ($B \times 1 \times d$) by design, as the satellite imagery with projected detections already captures spatial detail at multiple scales. Multi-token extraction from street images was considered but deemed redundant given this overlap, and would further increase the parameter count without commensurate gains. Exploring richer street-view representations (e.g., spatial token grids) remains a direction for future work. Overall, these results validate the multimodal design, where each modality contributes distinct and complementary semantic information that cannot be fully recovered from any single source.

\begin{table}[h]
\centering
\small
\caption{Ablation study on fusion, graph architecture, and message passing.}
\label{tab:ablation}
\resizebox{0.47\textwidth}{!}{%
\begin{tabular}{llc}
\toprule
\textbf{Component} & \textbf{Variant} & \textbf{P / R / F1 (\%)} \\
\midrule
\multirow{3}{*}{\shortstack{Fusion\\Strategy}}
& Early Fusion    & 41.2 / 38.1 / 39.5 \\
& Late Fusion     & 47.8 / \textbf{38.8} / 42.8 \\
& Cross-Attention & \textbf{58.0} / 37.5 / \textbf{45.5} \\
\midrule
\multirow{2}{*}{\shortstack{Graph\\Pooling}}
& Global          & 43.4 / \textbf{37.9} / 40.4 \\
& Edge-Aware      & \textbf{58.0} / 37.5 / \textbf{45.5} \\
\midrule
\multirow{2}{*}{\shortstack{Graph\\Encoder}}
& GAT             & 41.0 / 36.9 / 38.8 \\
& GATv2           & \textbf{58.0} / \textbf{37.5} / \textbf{45.5} \\
\midrule
\multirow{4}{*}{\shortstack{GNN\\Depth}}
& 1 Layer  & 41.4 / 37.4 / 39.3 \\
& 2 Layers & 53.6 / \textbf{38.4} / 44.8 \\
& 3 Layers & \textbf{58.0} / 37.5 / \textbf{45.5} \\
& 4 Layers & 53.4 / 37.4 / 43.9 \\
\midrule
\multirow{2}{*}{\shortstack{Message\\Passing}}
& Without & 43.7 / \textbf{38.0} / 40.7 \\
& With    & \textbf{58.0} / 37.5 / \textbf{45.5} \\
\bottomrule
\end{tabular}%
}
\end{table}

\paragraph{Fusion Strategy.}
Table~\ref{tab:ablation} compares alternative strategies for fusing graph and visual modalities. Early fusion, which concatenates features prior to processing, performs poorly ($F_1 = 39.5\%$), as it forces the model to learn complex cross-modal interactions without explicit structural guidance. Late fusion improves performance ($F_1 = 42.8\%$) by enabling modality-specific feature extraction, but still lacks fine-grained interaction during representation learning. In contrast, cross-modal attention achieves the best performance ($F_1 = 45.5\%$), providing a +2.7 improvement over late fusion. This confirms our design choice (Section~\ref{sec:fusion}) of allowing graph edges to query visual tokens, enabling selective alignment between candidate road segments and relevant visual evidence rather than relying on a fixed global fusion.

\paragraph{Graph Pooling Strategy.}
We compare global pooling, which averages over all $M$ edges including padding, with edge-aware pooling that aggregates only valid edges. Edge-aware pooling yields a substantial improvement of +5.1 $F_1$ points over global pooling (45.5\% vs.\ 40.4\%). This gain highlights the importance of explicitly handling variable-length graphs: global pooling dilutes informative representations by averaging over padded edges, particularly in sparse graphs, whereas edge-aware pooling ensures that only meaningful edge information contributes to the closure classification.

\paragraph{Graph Encoder Architecture.}
Table~\ref{tab:ablation} evaluates standard GAT against GATv2 for graph encoding. GATv2 significantly outperforms GAT (45.5\% vs.\ 38.8\%), demonstrating the benefit of more expressive, dynamic attention. By applying the nonlinearity after feature concatenation (Equation~\ref{eq:gatv2}), GATv2 enables flexible attention rankings over neighboring nodes. This expressiveness is critical for road closure reasoning, where the relevance of neighboring road segments varies with closure type, spatial context, and network topology.

\paragraph{GNN Depth.}
Table~\ref{tab:ablation} examines the effect of the number of message passing layers. A single GNN layer ($F_1 = 39.3\%$) captures only immediate neighborhood information, insufficient for understanding broader road network context. Two layers ($44.8\%$) enable 2-hop reasoning, substantially improving performance. Three layers achieve optimal results ($45.5\%$), allowing nodes to aggregate information from a sufficiently large neighborhood to understand local topology. Four layers ($43.9\%$) show slight degradation. This drop from 3 to 4 layers ($45.5\% \rightarrow 43.9\%$) is consistent with over-smoothing, a known phenomenon in deep GNNs where repeated message passing causes node representations to converge, reducing discriminability. Given the small graph sizes in our setting ($M{=}10$ edges), 3 hops already cover the full local neighborhood for most samples, making deeper propagation unnecessary.

\paragraph{Message Passing.}
Disabling message passing---using only initial node features without neighborhood aggregation---reduces $F_1$ by 4.8 points ($40.7\%$ vs.\ $45.5\%$). This demonstrates that topological context is crucial: message passing enables the model to reason about how roads connect and form network structure, and to distinguish between similar-looking roads based on their position in the network. Without message passing, the model relies solely on local edge and node attributes (distances, angles, road class). Enabling it allows each road segment to incorporate context from adjacent segments---for example, whether neighboring roads are also partially blocked, or whether the target segment connects to a dead end---which accounts for the observed improvement.

\paragraph{Class Imbalance.}
The natural class distribution in our dataset is imbalanced (62\% no closure, 29\% partial, 9\% complete), which disproportionately affects Complete Closure recall. With unweighted cross-entropy, per-class accuracy is 90.0\% for No Closure, 56.8\% for Partial Closure, and only 0.8\% for Complete Closure. To mitigate this, we train a variant using inverse-frequency weighted cross-entropy for the closure classification loss. Class weighting recovers Complete Closure accuracy from 0.8\% to 49.2\%, while No Closure decreases from 90.0\% to 73.9\% and Partial Closure from 56.8\% to 46.6\%. Overall, macro $F_1$ improves from 47.9\% to 53.0\% (+5.1) at the cost of reducing overall classification accuracy from 72.7\% to 63.8\% ($-8.9$ points). Segment association accuracy is largely unaffected (59.9\% $\rightarrow$ 58.7\%). These results confirm that the architecture can learn all closure types when given appropriate training signal, and that the tradeoff between overall accuracy and minority class recall is tunable via loss weighting without architectural changes. Further improvements through focal loss or targeted data augmentation for complete closures are noted as future work.

\paragraph{Error Analysis.}
We analyze model performance across road types, network complexity, and detection density to identify systematic failure patterns. For road segment association, performance degrades substantially as road hierarchy decreases: highways achieve 70.1\% accuracy compared to 54.1\% on arterials, 47.6\% on residential streets, and only 36.8\% on service roads. This reflects the increasing topological ambiguity in denser road networks where more candidate edges compete for selection. Network complexity, measured by the number of nodes in the local road graph, exhibits a similar effect: scenes with simple graphs ($\leq$4 nodes) achieve 72.2\% association accuracy versus 54.3\% for moderate-complexity graphs (5--7 nodes).

For closure classification, scenes with many detection events ($\geq$16) are harder to classify (69.4\%) compared to scenes with few detections (74.6\%, $\leq$5 events). Multiple detected objects introduce competing visual signals that increase classification ambiguity, consistent with the failure mode discussed in Section~\ref{sec:qualitative}. Conversely, association accuracy \emph{improves} with more detections (66.7\% vs.\ 54.7\%), as additional visual evidence provides stronger spatial cues for localizing the affected road segment. These opposing trends suggest that detection multiplicity benefits spatial reasoning while complicating closure type inference, motivating future work on multi-object disambiguation.

\subsection{Qualitative Analysis}
\label{sec:qualitative}

\begin{figure}[ht]
  \centering
  \begin{subfigure}{0.55\linewidth}
    \centering
    \includegraphics[width=\linewidth]{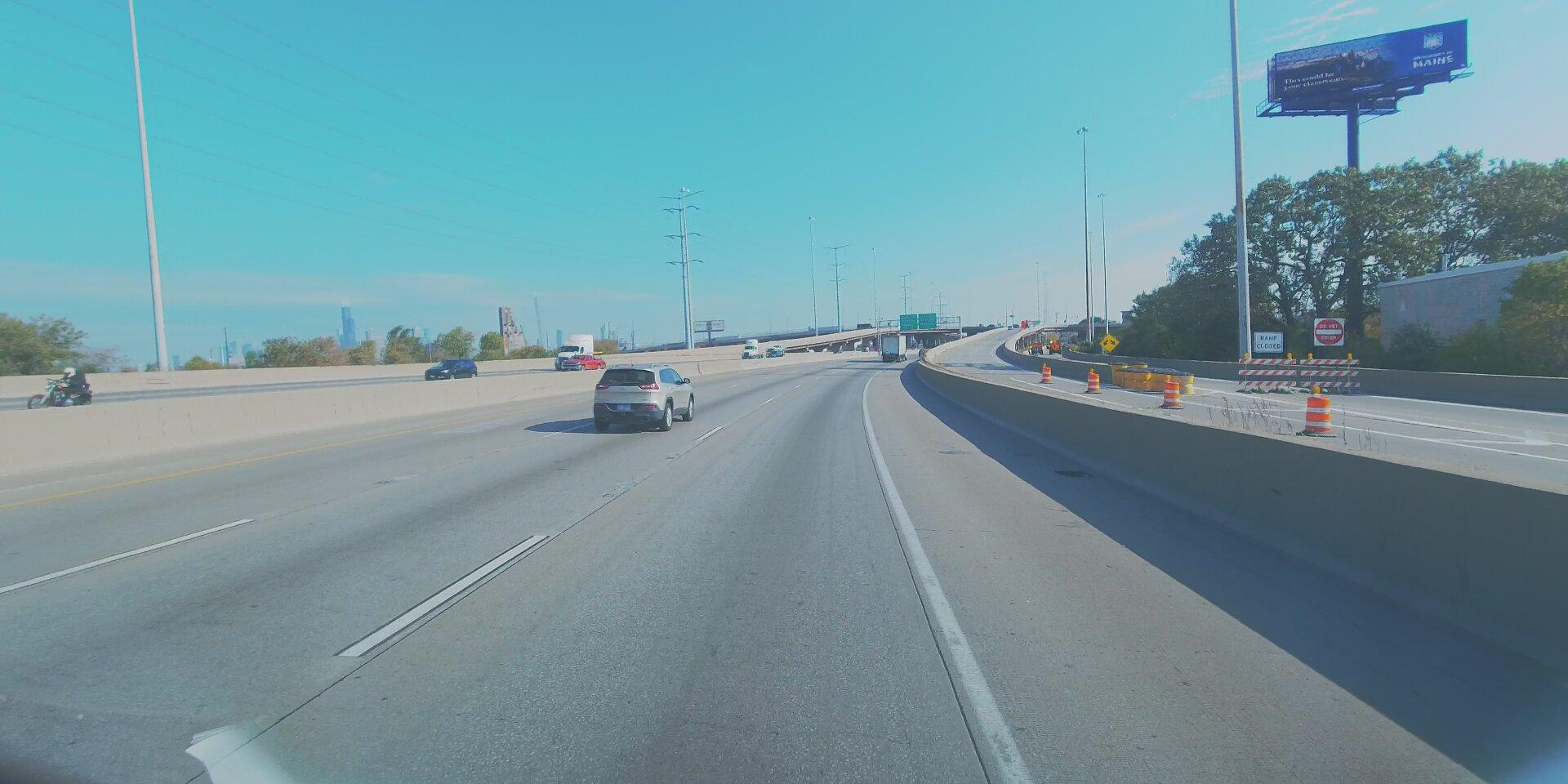}
    \caption{Street imagery for a "Complete" closure sample.}
  \end{subfigure}
  \hfill
  \begin{subfigure}{0.44\linewidth}
    \centering
    \includegraphics[width=\linewidth]{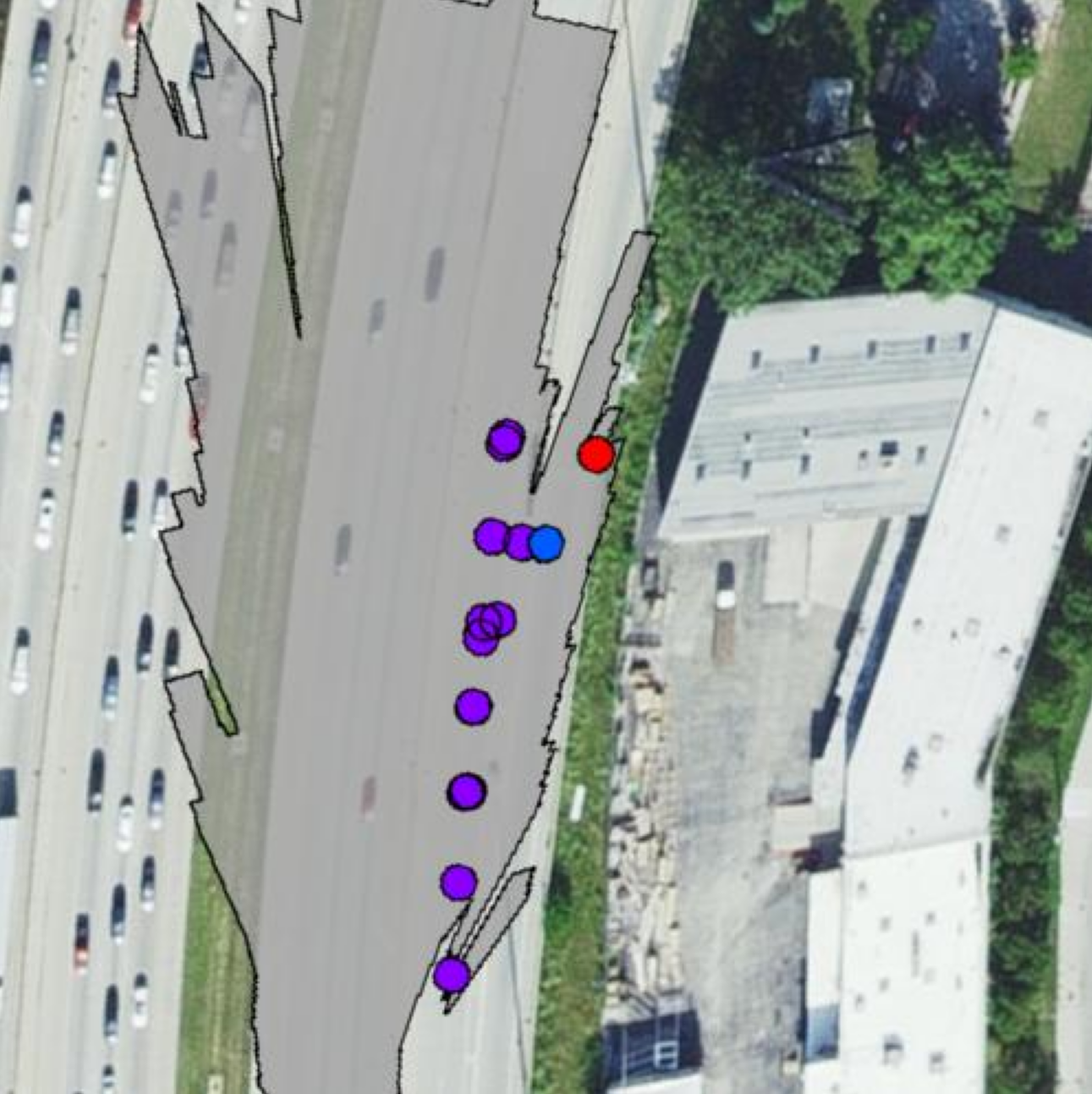}
    \caption{Satellite imagery for a "Complete" closure sample.}
  \end{subfigure}
  \\
  \begin{subfigure}{0.55\linewidth}
    \centering
    \includegraphics[width=\linewidth]{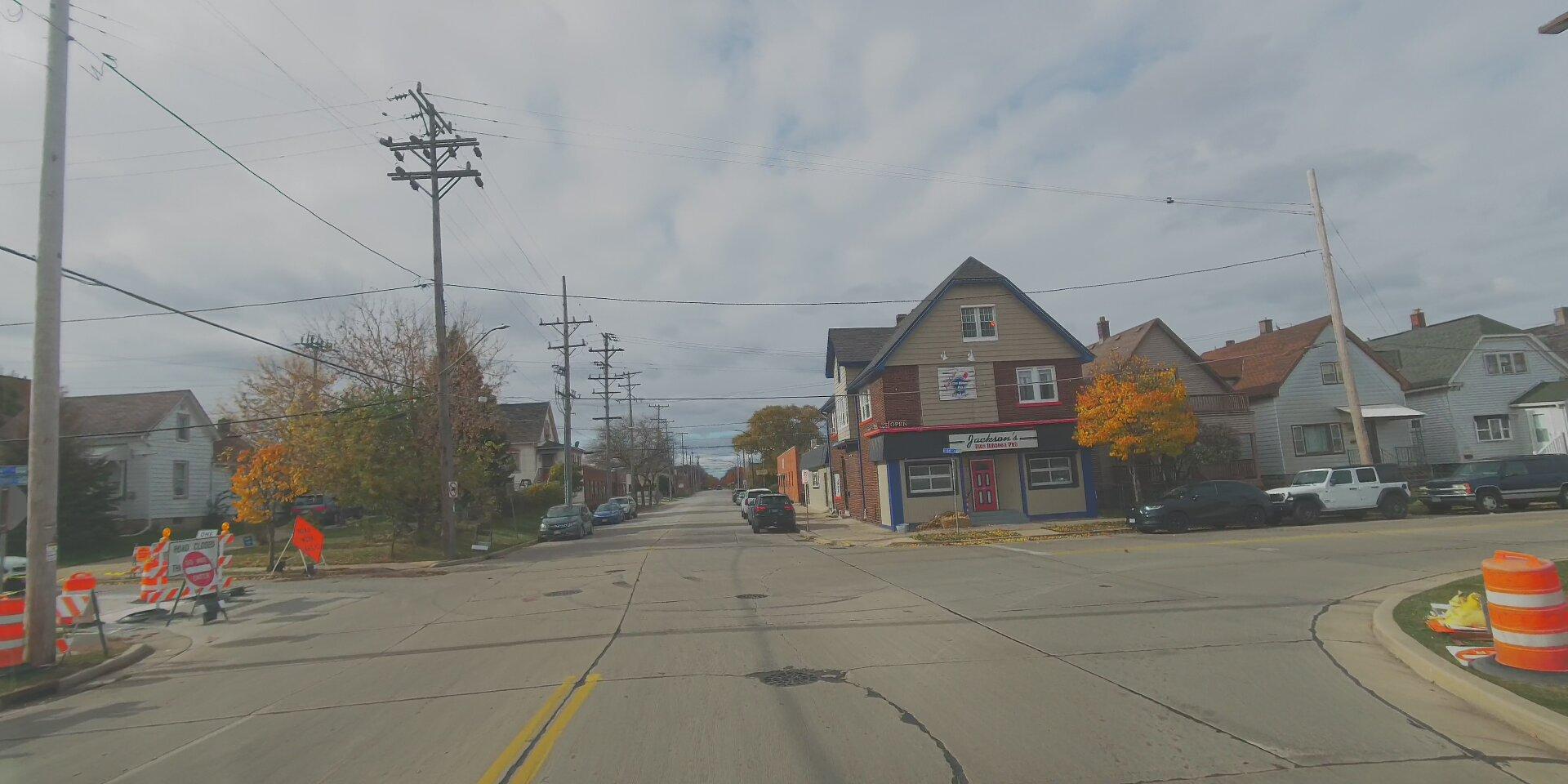}
    \caption{Street imagery for a "Partial" closure sample.}
  \end{subfigure}
  \hfill
  \begin{subfigure}{0.44\linewidth}
    \centering
    \includegraphics[width=\linewidth]{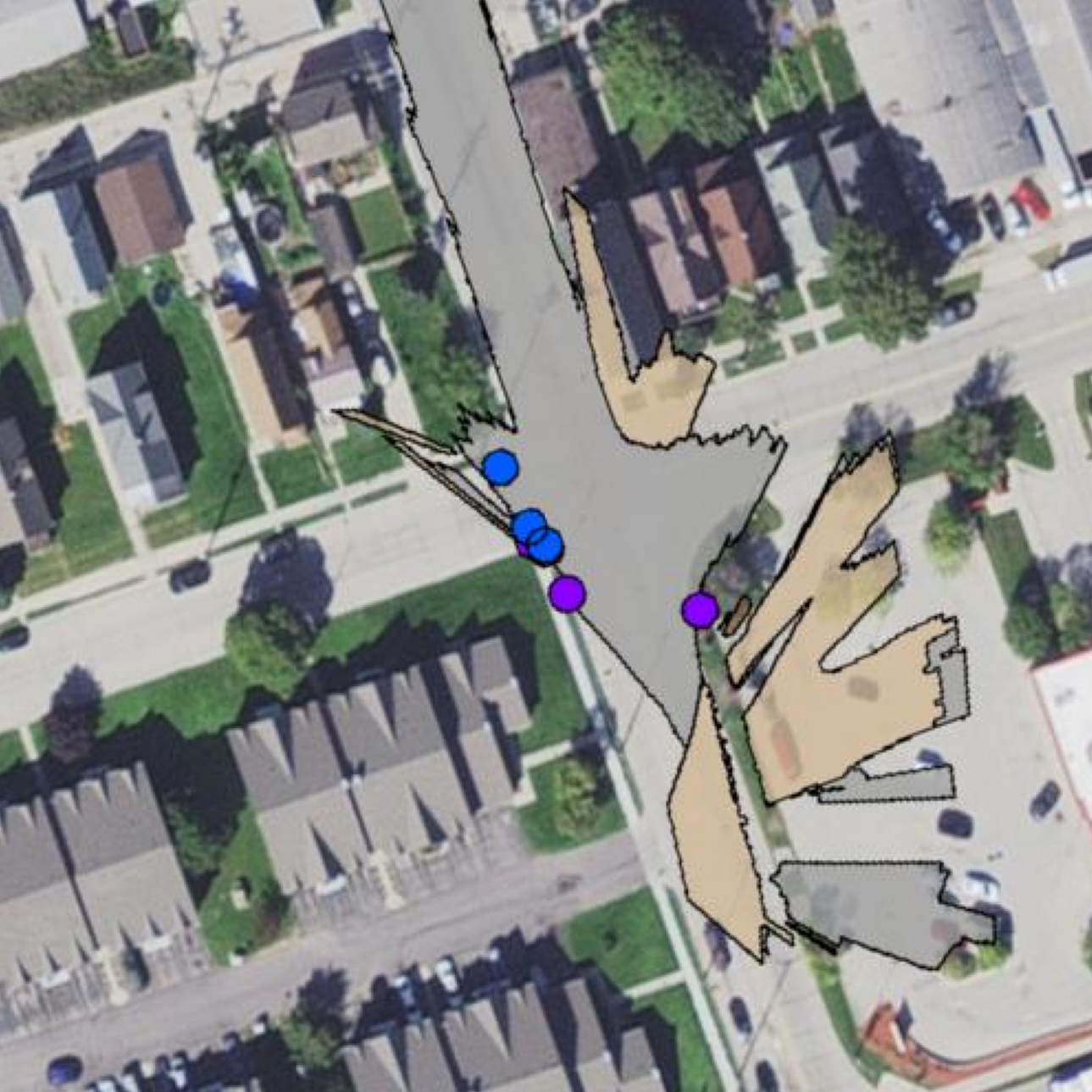}
    \caption{Satellite imagery for a "Partial" closure sample.}
  \end{subfigure}
  \caption{Two successful examples for partial and complete closures.}
  \label{fig:qual_success}
\end{figure}

Figure~\ref{fig:qual_success} presents qualitative examples of successful predictions. Our model correctly identifies complete closures on major highways (row 1) and partial closures on residential streets (row 2), demonstrating its ability to reason across diverse road types and closure scenarios.

\begin{figure}[ht]
  \centering
  \begin{subfigure}{0.8\linewidth}
    \centering
    \includegraphics[width=\linewidth]{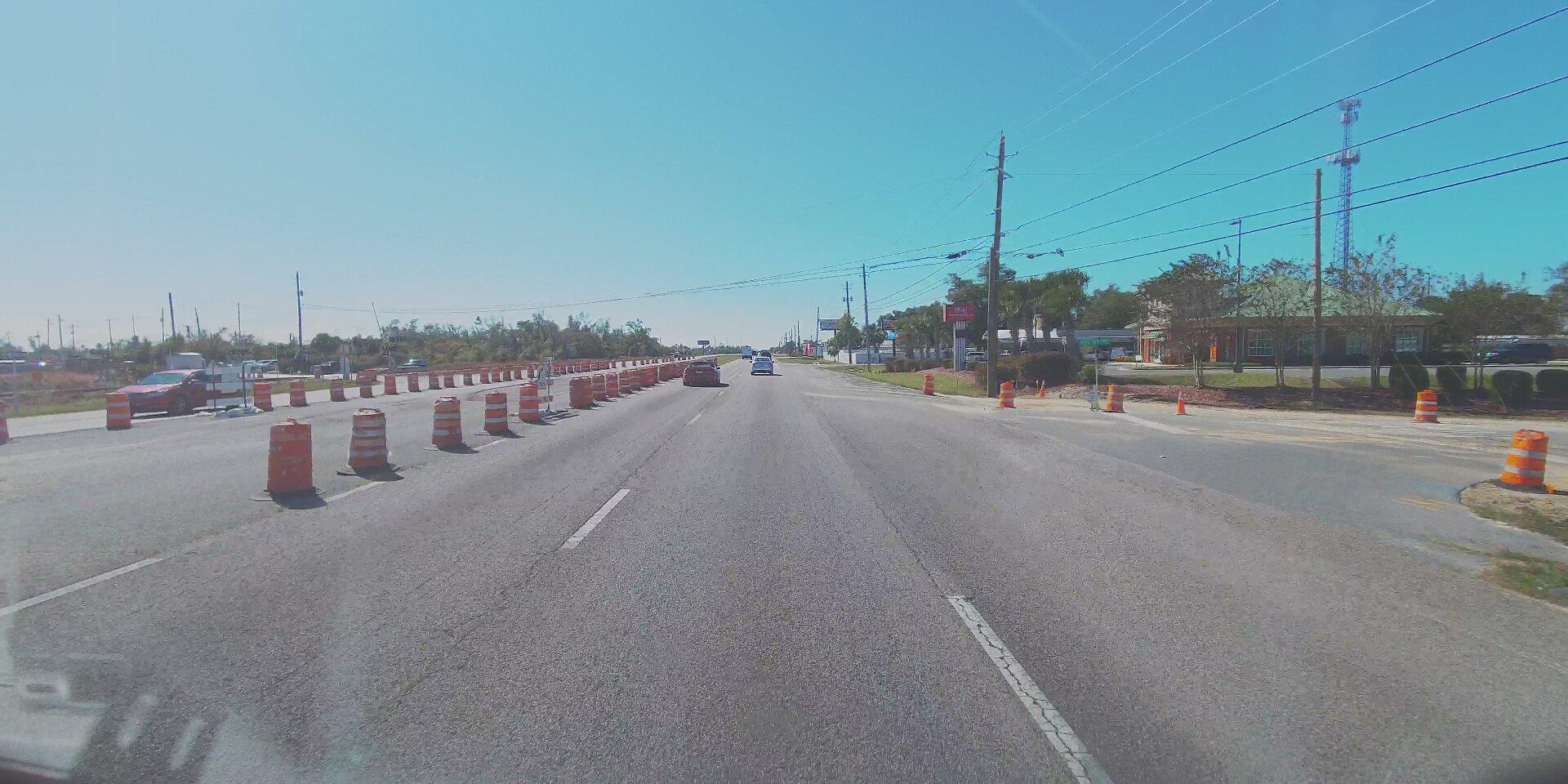}
    \caption{Street imagery.}
  \end{subfigure}
  \vfill
  \begin{subfigure}{0.48\linewidth}
    \centering
    \includegraphics[width=\linewidth]{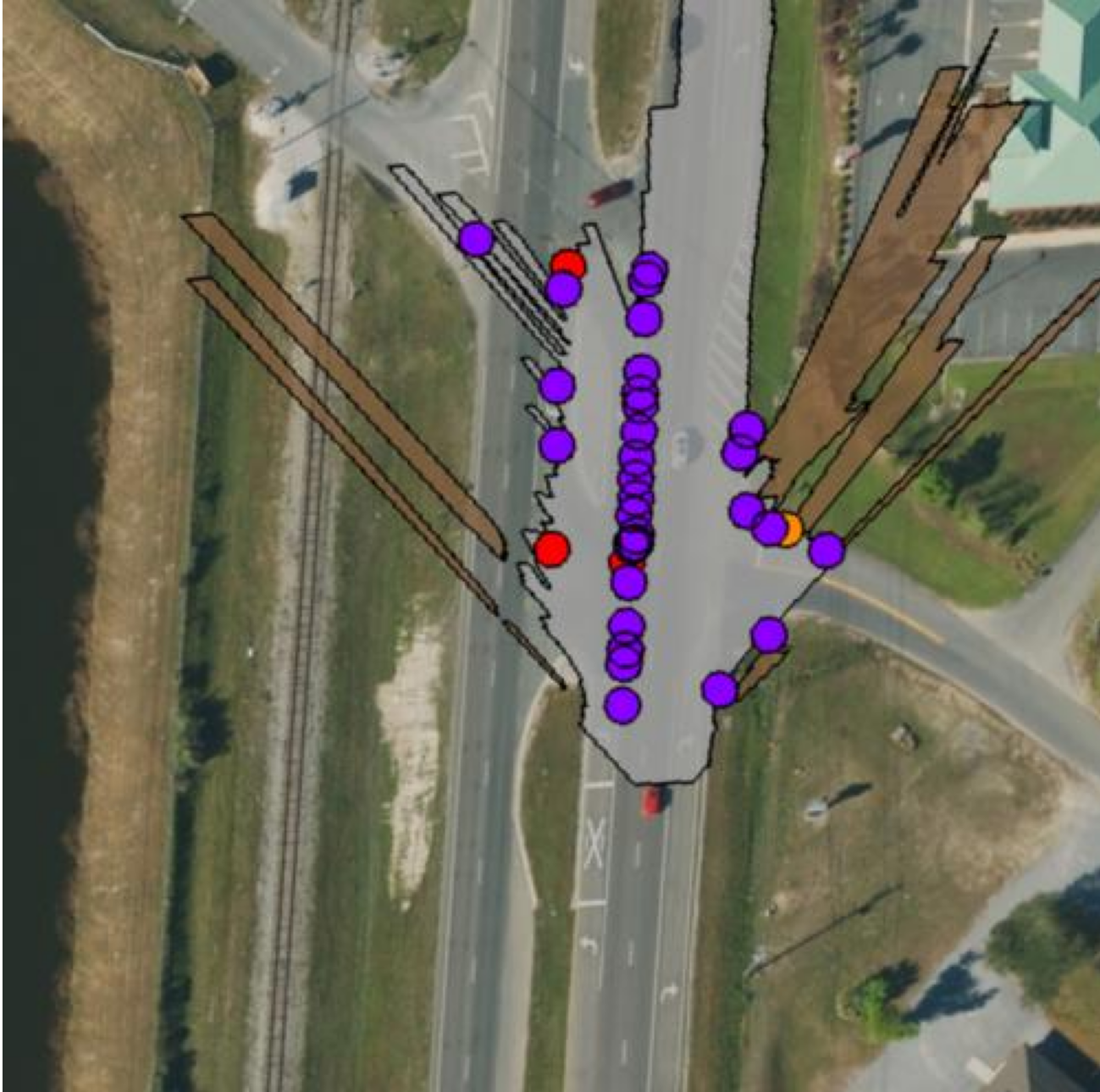}
    \caption{Projected closure objects.}
  \end{subfigure}
  \hfill
  \begin{subfigure}{0.48\linewidth}
    \centering
    \includegraphics[width=\linewidth]{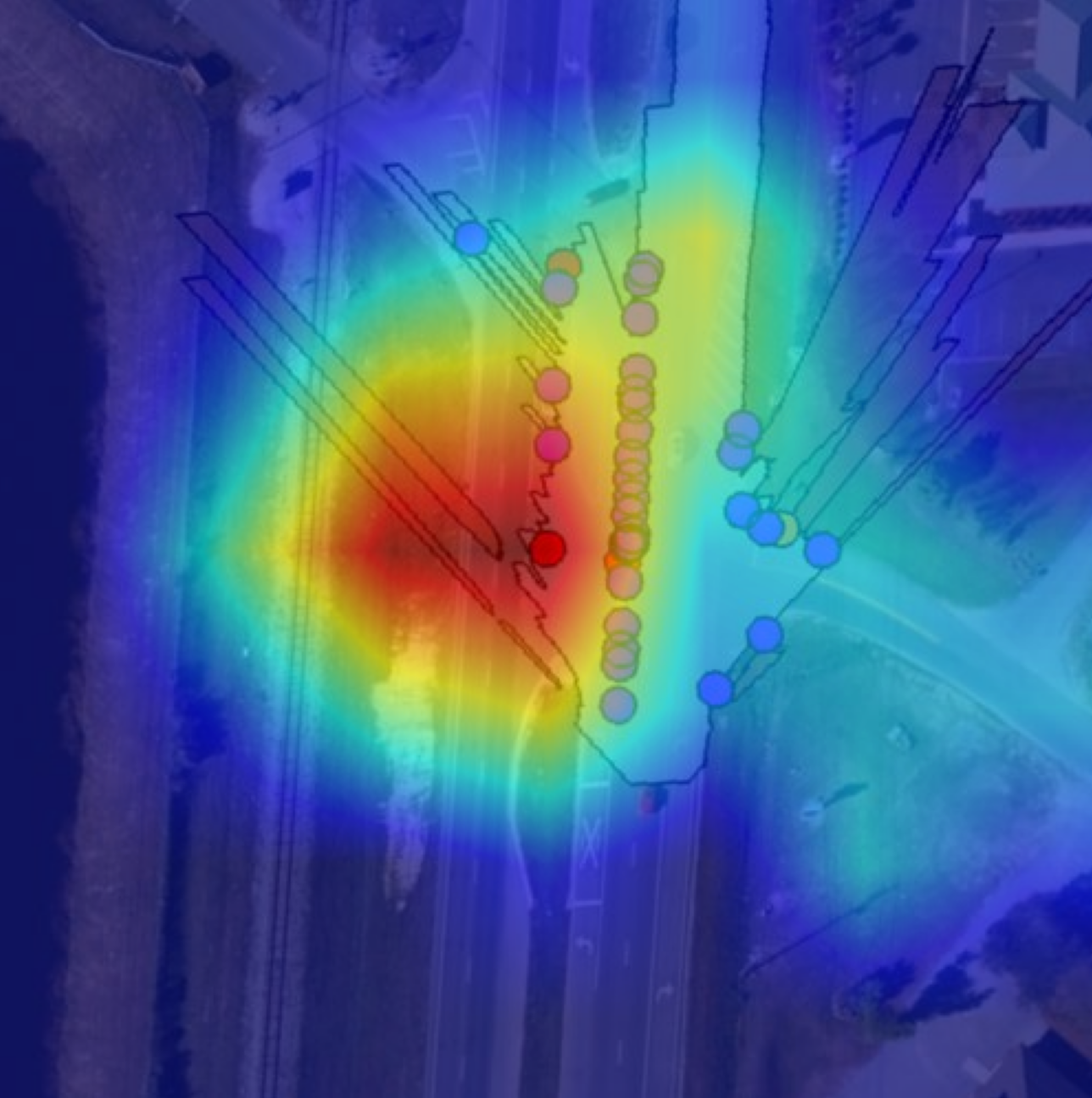}
    \caption{Model activation visualization.}
  \end{subfigure}
  \caption{Anecdote showing an ambiguous sample.}
  \label{fig:anecdote}
\end{figure}

An illustrative failure case highlighting the task’s complexity is shown in Figure~\ref{fig:anecdote}. The scene contains a visually prominent line of drums, which the model has learned to associate with partial closures. Consequently, it attends to the adjacent road segment and predicts a partial closure. However, the decisive cue is a barely visible barricade located between the drums, indicating a complete closure on a different road segment. The model fails to capture this subtle signal and is instead dominated by the more salient drum pattern. This example exposes two challenges: strong learned correlations between object types and closure categories (drums $\rightarrow$ partial closure), and the difficulty of reasoning when multiple closure objects with different implications are present, where the model may focus on the more prominent ones while missing subtle but decisive indicators. These observations motivate future work on detecting and associating multiple closures within a single scene, beyond the current single-closure-per-sample assumption, as well as more expressive attention mechanisms to disambiguate competing visual evidence.

\section{Conclusion}
\label{sec:conclusion}

We presented a multimodal, multitask framework for mapping dynamic road closures. Our architecture encodes street-level imagery, satellite imagery with projected detections, and graph-structured road networks through cross-modal attention, enabling joint reasoning over visual cues and spatial topology. The proposed model substantially outperforms general-purpose vision--language model baselines while being significantly smaller and faster. This work establishes a benchmark for joint closure classification and road segment association, demonstrating that task-specific architectural inductive biases are essential for structured geospatial reasoning. The framework generalizes to other road attributes such as lane restrictions and turn prohibitions.

\paragraph{Future Work.}
Future directions include inferring closure end times from temporal visual cues, handling multiple simultaneous closures within a single scene, and evaluating the framework in operational settings at scale.


{\small
\bibliographystyle{ACM-Reference-Format}
\bibliography{ref}
}

\clearpage
\appendix
\section{Hyperparameters and Model Specifications}
\label{appendix:hyperparameters}

This appendix provides detailed hyperparameters and architectural specifications for the proposed multimodal model.

\subsection{Model Architecture Hyperparameters}

\begin{table}[ht]
\centering
\caption{Model architecture hyperparameters and specifications.}
\label{tab:model_hyperparameters}
\resizebox{0.47\textwidth}{!}{%
\begin{tabular}{lll}
\toprule
\textbf{Component} & \textbf{Parameter} & \textbf{Value} \\
\midrule
\multicolumn{3}{l}{\textit{Global Settings}} \\
& Hidden dimension ($d$) & 512 \\
& Dropout rate & 0.2 \\
\midrule
\multicolumn{3}{l}{\textit{Graph Encoder (GATv2)}} \\
& Number of GAT layers & 3 \\
& Number of attention heads & 8 \\
& Node feature dimension & 9 \\
& Edge feature dimension & 9 \\
& Maximum edges per graph ($M$) & 10 \\
& Use edge bootstrap & True \\
& Use message passing & True \\
\midrule
\multicolumn{3}{l}{\textit{Street Image Encoder}} \\
& Backbone architecture & ResNet-18 \\
& Pretrained weights & ImageNet-1K \\
& Input resolution ($H_{st} \times W_{st}$) & $224 \times 224$ \\
& Output dimension & $d$ (512) \\
\midrule
\multicolumn{3}{l}{\textit{Aerial Image Encoder (FPN)}} \\
& Backbone architecture & ResNet-18 \\
& Pretrained weights & EuroSAT \\
& Input resolution ($H_s \times W_s$) & $224 \times 224$ \\
& FPN output channels & $d$ (512) \\
\midrule
\multicolumn{3}{l}{\textit{Cross-Modal Fusion}} \\
& Fusion strategy & Cross-attention \\
& Number of transformer layers ($N_{\text{layers}}$) & 2 \\
& Number of attention heads ($H$) & 8 \\
& Dropout & 0.2 \\
\bottomrule
\end{tabular}
}
\end{table}

\subsection{Training Hyperparameters}

\begin{table}[H]
\centering
\caption{Training hyperparameters and optimization settings.}
\label{tab:training_hyperparameters}
\begin{tabular}{ll}
\toprule
\textbf{Parameter} & \textbf{Value} \\
\midrule
Optimizer & AdamW \\
Learning rate & $1 \times 10^{-4}$ \\
Weight decay & $1 \times 10^{-4}$ \\
Batch size & 32 \\
Number of epochs & 60 \\
Learning rate scheduler & ReduceLROnPlateau \\
Scheduler patience & 5 epochs \\
Scheduler factor & 0.5 \\
Loss weights ($\lambda_{\text{assoc}}$, $\lambda_{\text{closure}}$) & 1.0, 1.0 \\
\bottomrule
\end{tabular}
\end{table}

\subsection{Data Preprocessing}

\begin{table}[H]
\centering
\caption{Data preprocessing and augmentation specifications.}
\label{tab:preprocessing}
\begin{tabular}{ll}
\toprule
\textbf{Parameter} & \textbf{Value/Method} \\
\midrule
\multicolumn{2}{l}{\textit{Image Normalization}} \\
Mean (RGB) & [0.485, 0.456, 0.406] \\
Std (RGB) & [0.229, 0.224, 0.225] \\
\midrule
\multicolumn{2}{l}{\textit{Graph Features}} \\
Road class normalization & Ordinal (10-102 $\rightarrow$ 0-1) \\
Distance normalization & Min-max \\
Angle normalization & Sin and Cos \\
\midrule
\multicolumn{2}{l}{\textit{Aerial Image Processing}} \\
Satellite image coverage & $100m \times 100m$ \\

\bottomrule
\end{tabular}
\end{table}

\subsection{Computational Resources}

\begin{table}[H]
\centering
\caption{Computational resources and training time.}
\label{tab:computational}
\begin{tabular}{ll}
\toprule
\textbf{Resource} & \textbf{Specification} \\
\midrule
GPU & Tesla V100 (16GB) \\
Number of GPUs & 1 \\
Training time per epoch & $\sim$4 minutes \\
Total training time & $\sim$4 hours \\
\bottomrule
\end{tabular}
\end{table}

\end{document}